\RequirePackage{fix-cm}
\documentclass{svjour3}                     
\smartqed  
\usepackage{graphicx}

\makeatletter
\let\cl@chapter\undefined
\makeatletter
\usepackage[T1]{fontenc}
\usepackage{natbib}
\usepackage{graphicx}
\usepackage{amssymb}
\usepackage{mathrsfs}
\usepackage{amsmath}
\usepackage{float}
\usepackage{subfig}
\usepackage{booktabs}
\usepackage{url}
\usepackage{diagbox}
\usepackage[table]{xcolor}
\usepackage{multirow}
\usepackage{tabularx}
\newcolumntype{b}{X}
\newcolumntype{m}{>{\hsize=.6\hsize}X}
\newcolumntype{s}{>{\hsize=.25\hsize}X}
\newcommand{\heading}[1]{\multicolumn{1}{c}{#1}}

\usepackage{amssymb}
\usepackage{amsmath}
\usepackage{cleveref}
\begin{document}

\title{Machine learning for assessing quality of service in the hospitality sector based on customer reviews
}

\titlerunning{ML \& quality of service in hospitality customer reviews}        
\author{Vladimir Vargas-Calderón \and {Andreina~Moros~Ochoa} \and {Gilmer~Yovani~Castro~Nieto} \and {Jorge~E.~Camargo}}
\institute{Vladimir Vargas-Calderón \at Laboratorios de Investigación en Inteligencia Artificial y Computación de Alto Desempeño, Human Brain Technologies, Bogotá, Colombia.  \at Grupo de Superconductividad y Nanotecnología, Departamento de Física, Univesidad Nacional de Colombia, Bogotá, Colombia. \email{vvargasc@unal.edu.co} \and Andreina Moros Ochoa \at Colegio de Estudios Superiores de Administración, Diagonal 35 \#5a-57, Bogotá, Colombia. \and Gilmer Yovani Castro Nieto \at Departamento de Administración de Empresas, Pontificia Universidad Javeriana, Carrera 7 No. 40B-36, Bogotá, Colombia. \and Jorge E. Camargo \at System Engineering Department, Fundación Universitaria Konrad Lorenz, Bogotá, Colombia.}




\date{ }

\maketitle

\begin{abstract}

The increasing use of online hospitality platforms provides firsthand information about clients preferences, which are essential to improve hotel services and increase the quality of service perception. Customer reviews can be used to automatically extract the most relevant aspects of the quality of service for hospitality clientele. This paper proposes a framework for the assessment of the quality of service in the hospitality sector based on the exploitation of customer reviews through natural language processing and machine learning methods. The proposed framework automatically discovers the quality of service aspects relevant to hotel customers. Hotel reviews from Bogotá and Madrid are automatically scrapped from Booking.com. Semantic information is inferred through Latent Dirichlet Allocation and FastText, which allow representing text reviews as vectors. A dimensionality reduction technique is applied to visualise and interpret large amounts of customer reviews. Visualisations of the most important quality of service aspects are generated, allowing to qualitatively and quantitatively assess the quality of service. Results show that it is possible to automatically extract the main quality of service aspects perceived by customers from large customer review datasets. These findings could be used by hospitality managers to understand clients better and to improve the quality of service.

\end{abstract}

\section{Introduction}

Quality of service is a fundamental element in the hospitality industry, since delivering high quality of service is a differential feature that positively impacts hospitality businesses, particularly on their performance and competitive position~\citep{wong1999analysing}. The quality of service is a key ingredient that makes hospitality businesses thrive because there is a direct link between the quality of service and customer satisfaction, which translates into loyalty and repurchasing intention~\citep{anderson1993,ghotbabadi2012review}. Accordingly, the need for measurement of quality of service in the hospitality sector has been identified as critical in the last few decades because measuring it allows businesses to set goals related to the improvement of profitability, keeping track of their fulfilment.

In this order of ideas, frameworks to measure the quality of service have been built responding to different measuring needs. Researchers and managers usually disregard standardised frameworks since not every hospitality business thrives for the same population niche, meaning that failing to achieve a gold standard does not imply worse business performance. Instead, customer-oriented frameworks are preferred, as they directly correlate with customer satisfaction, which is related to the achievement of the customer's expectations~\citep{anderson1993}.

As might be expected, researchers and managers have produced a plethora of various methods to measure the quality of service, which have been applied to the hospitality sector~\citep{ghotbabadi2012review,lai2018literature}. The process of obtaining and gathering information from the clients has generally been done through surveys, which are time- and money-consuming. However, the growth of online hospitality platforms has generated firsthand data about customers that is not being used by these conventional methods. Understanding this data and finding ways to improve how customers perceive the quality of service is utterly essential, primarily because travellers use to write and read online reviews at any stage of their travel to make purchase decisions~\citep{zhou2020tourists,pourfakhimi2020electronic}. 

The development of these technological platforms enables the possibility to share opinions and assessments between users of services or products, generating a large volume of valuable information for the sector, which researchers can access to improve quality of service models~\citep{XIANG201751}. Through the Internet, consumers actively use the dissemination of information by engaging in social dynamics where they receive and emit individual interactions, comments and decisions. This engagement is also known as the electronic word of mouth (eWOM) \citep{williams2019wom,pourfakhimi2020electronic}. In fact, it has been shown that potential tourists are more likely to seek information about a destination through messages or comments from other consumers~\citep{abubakar2017ewom} in virtual platforms to make purchasing decisions~\citep{MATUTEVALLEJO201561,septianto2018effects,kim2018influence}. Therefore, eWOM is important because it improves tourists' experiences before, during, and after the trip~\citep{rahmani2018tourists}. Those customer opinions and assessments about the quality of service are user gathered content that is valuable for managers to understand customers better and modify internal processes to improve user satisfaction. 

The amount of information produced in these platforms sets a Big Data scenario, which --in recent years-- has played a central role in tourism and hospitality research and delineates how it might evolve in the future \citep{mariani2018business,mariani2019big}. The use of Big Data might allow making strategic decisions using current customer data \citep{lamest2019data} and improve the competitiveness of tourism organisations and destinations \citep{buhalis2019technology}. Most of this Big Data is expressed in text or images (customer reviews). It has been proposed that such Big Data can be exploited through artificial intelligence methods~\citep{buhalis2019technological} able to process text~\citep{xiang2019online,ma2018sentiment} and images~\citep{MA2018120} to assess several aspects of the hospitality industry. This is of particular usefulness, taking into account other sources different from online hospitality platforms such as social media~\citep{moro2019evaluating,lin2020task}.

The availability of this Big Data is crucial, especially to better assess the quality of service in the hospitality sector using customer reviews~\citep{rahmani2019psycholinguistic}. Therefore, the ability to effectively analyse data, using, in occasions dedicated software, becomes a crucial aspect of hotel management. Consequently, it is natural to ask ourselves if we can measure and identify the main topics of the perceived quality of service in hospitality businesses from non-structured data contained in online customer reviews. In principle, the answer is yes: this is exactly what quality of service methods do. However, these methods would require investing human resources to extract meaningful and measurable statistics from the reviews so that the retrieved information can fit the dimensions and variables typically measured by the conventional quality of service measurement methods. In this work, we propose an alternative to assess the quality of service in the hospitality sector through the use of Natural Language Processing (NLP), a branch of artificial intelligence, to tackle the problem of inferring meaningful and measurable information from large customer opinions datasets that can be gathered from online hospitality platforms.

Specifically, we propose an NLP framework for rigorously studying customer opinions datasets, whose contribution is two-fold. First, our framework helps in qualitatively assessing the quality of service, focusing on and highlighting the main topics that concern the customers and the topics that comfort them. The fine-grained topic information can be obtained at different levels: city-, location- and business-level. Second, we provide an analysis of the identified topics and their relation to customers' final scores, which offers a quantitative measure of the quality of service for each topic.

We exemplify our NLP framework by obtaining city-level information about concern and comfort topics from hospitality customers in Bogotá, Colombia, and Madrid, Spain, through online reviews written in Spanish found at Booking.com for any hospitality business. We decided to study these two cities because they are capitals that generally have significant tourist traffic. \citet{DHAR2015419} highlights the importance of tourism development in emerging economies, as is the case of Colombia. \citet{citur2020} claims that in recent years, foreigner arrivals to the country have grown exponentially: in 2018, more than 4.2 million non-resident visitors, 8,504,778 travellers arrived in Bogotá on national flights and 4,465,741 on international flights; hotel occupancy in the country was 56.7\%. For its part, Spain ranks third in the EU with the highest number of international tourists, 83.7 million non-resident travellers visited the country in 2019. 87.4\% of these visits were for leisure, recreation and vacations and 6.4\% for business and professional reasons~\citep{ine2020}. Specifically, Madrid recorded 10.4 million visitors in 2018~\citep{ine2020} and the country registered a hotel occupancy rate of more than 73\%.

This paper is divided as follows. \Cref{sec:overview} shows an overview of machine learning applications in the hospitality sector. In~\cref{sec:methods} we expose a deeper review of the methods used for text analysis in the hospitality sector. We also describe the data set used for this study and explain in detail the design of our research. Later, in~\cref{sec:results} we present the results of this work and finally conclude in~\cref{sec:conclusions}.

\section{Literature review}
\label{sec:overview}

Some of the main historical contributions to the quality of service assessment are the model proposed by \citet{gronroos1984service}, the widely used Servqual \citep{parasuraman1994reassessment}, the Servperf model~\citep{cronin1992measuring}, the retail service quality scale~\citep{dabholkar1996measure} and the hierarchical and multidimensional model for service quality~\citep{brady2001some}. However, the Servqual model has been the most cited one among researchers. It defines five service quality dimensions: tangible elements, reliability, responsiveness, empathy and assurance, in turn, composed of 22 variables~\citep{parasuraman1994reassessment,parasuraman1988servqual}. As a matter of fact, some authors have made adaptations of the model for measuring quality specifically in hotels such as Lodgserv \citep{knutson1990lodgserv}, Holserv \citep{wong1999analysing}, Hotelqual  \citep{hernandezcalidad},Resortqual \citep{alen_2004}; Caltic \citep{ochoa2016adaptation}, SMSHs \citep{ahmad2018measuring}, Glserv \citep{lee2018less}, and models without a proper name such as Servqual with 29 variables \citep{akbaba2006measuring} and the same five dimensions (Tangibles, Reliability, Responsiveness, Assurance and Empathy) and twenty-two variables \citep{lestari2018service,lestari2018market,keshavarz2018service}. 

As with every quality of service model, Servqual faces the task of gathering information from customers to assess their perception of the quality of service. Usually, surveys are designed and applied to retrieve this information. However, the amount of data that can be recovered with surveys can be limited by time and financial budgets. Therefore, using massive datasets from online hospitality platforms offers an exciting and valuable alternative to retrieve information, with the hindsight of being non-structured. As opposed to well-controlled surveys, information found in hospitality platforms is much more diverse because customers express their opinions with complete freedom~\citep{zhou2020tourists,pourfakhimi2020electronic}. Therefore, methodologies must be devised to process and extract useful information from the data found ins online hospitality platforms. In this regard, opinion mining and machine learning methods have provided several good results in numerous contexts.

Machine learning is an area of computer science related to the understanding and design of algorithms that solve performance-measurable tasks which automatically improve through experience~\citep{mitchell1997machine}. The tasks intended to be solved through machine learning cover the whole complexity spectrum and reside in many fields of human knowledge. There are two branches of machine learning which are relevant for this work: supervised and unsupervised learning. In supervised learning, algorithms are expected to discover a rule set that produces an expected output given an input. The expected output is known \textit{a priori}, as it is usually the case that one has a training set of examples with pairs of input-output. To illustrate, consider a dataset where one has newspaper headlines as inputs and news categories as outputs. The reason for discovering a rule set that matches inputs with outputs might be to build an automatic classifier that takes any future newspaper headline and categorises it, giving it an output. Thus, the learning is said to be supervised, as one wants the algorithm to maximise the number of correctly classified headlines.

Regarding unsupervised learning, algorithms are expected to discover patterns from a collection of objects. If we stick to the same illustration, one can have a dataset with only newspaper headlines and no categories. A question that may emerge is, can we create groups of newspaper headlines without \textit{a priori} knowledge of the actual categories? Unsupervised learning handles these sorts of tasks. In both cases, the more data there is, the better the algorithms usually perform: having more data is related to having more ``experience''.

In a more broad scenario, machine learning is considered an area within artificial intelligence. The impact of the latter on the hospitality industry has been more studied so far. Mainly, in recent years, researchers have devoted efforts to the following issues: understanding and measuring the possible acceptance of human-robot interaction in the hospitality sector~\citep{lin2019antecedents,John2018}; producing forecasting methods to timely anticipate the demand~\citep{onder2019forecasting}; reviewing technology development of possible artificial intelligence-related products in hospitality~\citep{chi2020artificially}, and measuring consumer experience~\citep{Sun2018}. The last research branch has produced only a limited amount of research, mainly because in order to develop work in this area, a high interdisciplinary collaboration is needed between researchers of hospitality and tourism and researchers of mathematical methods in machine learning with the ability to gather large datasets to be analysed. Such is the case of our work.

In the quest to use the vast amount of data produced by consumers of hospitality services, machine learning and big data analytics are highlighted as promising tools to unveil hidden patterns that are important to discern consumer dynamics and study them rigorously. Machine learning has recently started to be used in this area with success. For instance, \citet{lee2018assessing} collected around a million hotel reviews and developed effective classification models to assess review helpfulness by using data mining techniques. Also, \citet{martin2018modelling} created a model to classify the properties offered by peer-to-peer (P2P) accommodation platforms, similar to grading scheme categories of hotels, for predicting a hotel category by taking into consideration certain quality variables. This model was applied based on information extracted from 18 million reviews from Booking.com and applied to Airbnb to predict their star rating. \citet{AHANI201952} used clustering techniques and dimensionality reduction methods to predict users interests for market segmentation purposes in Spa-hotels. Other machine learning techniques such as recurrent neural networks, long short term memory networks and convolutional networks were jointly used by \citet{MA2018120} to extract features from text and images to study how these affect hotel review helpfulness. Even basic yet powerful methods based on word frequency analysis have been used to determine the major drivers of customers reviews~\citep{PADMA2020102318}.


Moreover, \citet{CHENG201958} used Leximancer~\citep{Smith2006} to characterise the main attributes taken into account by hotel clients when they review hotels. In the same direction, \citet{LUO2019144} used a modified version of latent aspect rating analysis to study critical features described by users when expressing their experience at a hotel. Despite the recent use of machine learning techniques in the hospitality sector, these studies do not address the quality of service measurement as their central task. However, we must emphasise that the study of comments has been proposed as an essential tool at high-level decision-making instances in other domains because it captures plenty of information about what people think or how they feel about something (e.g. \citep{chen2018research,agrawal2018prediction,luo2019research,TAECHARUNGROJ2019550,MARTINEZTORRES2019393}).

Therefore, research using machine learning methods for processing online customer reviews in the hospitality sector has started to flourish recently. Still, none has focused on assessing the quality of service of the hospitality sector. Our work aims to fill this knowledge gap and to start a discussion on how to integrate robust and characterised methods such as Servqual with these vast sources of information.

\section{Research Methods}
\label{sec:methods}

In this section, we qualitatively review some of the machine learning methods used in the area of NLP, which are needed to automatically process and draw valuable information from large sets of texts. We do so by paying closer attention to the use of such NLP methods in the related work. Furthermore, we also discuss the elements from these works that are useful for our research which has the objective of qualitatively and quantitatively assess the quality of service in the hospitality sector. Here, the quality of service is understood from the point of view of~\citet{parasuraman1994reassessment} in their Servqual model, where it can be assessed by measuring the gap between the expectation and perception of service from the customers' opinions. However, we must emphasise that our approach to measuring this gap depends on the assumption that customers who post reviews on online hospitality platforms do so mainly by highlighting the experiences that either surpassed or failed to meet their expectations. In that sense, the objective of our study is to propose and test a methodology to survey the topics that concern or satisfy the customers the most and that can identify the level of satisfaction-dissatisfaction of a customer. The last point is crucial, as we are mainly interested in quantifying which aspects of the quality of service affect the most the customers' perception.

As a result of the main objective so far exposed, NLP techniques allow us to process the massive amounts of text encountered in online hospitality platforms. The backbone of the methodology that we propose in this work is familiar to many NLP applications: the text needs to be numerically represented to conduct clustering tasks or conduct prediction tasks, among other machine learning tasks. There are several alternatives to numerically represent text, grouped into feature-engineering methods and unsupervised methods. The former refers to using human and context knowledge to extract features that resemble and exhibit the semantics of text. The latter refers to language models that exploit the co-occurrence of words in different sentences to infer semantic structure in text.

An example of feature-engineering can be found in the work by~\citet{lee2018assessing}, where a model for predicting the helpfulness of reviews in TripAdvisor was built. They extracted some text features from the reviews in order to numerically represent it. Some of the features are the number of characters, words, syllables and sentences, as well as some readability indices. Moreover, already implemented machine learning models (particularly Stanford's CoreNLP~\citep{manning-EtAl:2014:P14-5}) were used to extract more sophisticated features such as the sentiment of the reviews. Other variables, such as the reviewer age or gender, were also used. Therefore, each customer review was represented as a vector, where each component held one of the engineered features. Those vectors were subsequently used to predict each review's helpfulness, which is defined as a rate of helpful votes from other TripAdvisor users. The authors tested several classification algorithms aiming to classify a review as helpful or not. Among the used algorithms were decision trees, random forests, logistic regression and support vector machines. Consistently, random forest was the most accurate algorithm in performing the classification.

Another closely related study to predict the helpfulness of reviews from TripAdvisor and Yelp is conducted by~\citet{MA2018120}. However, this is an example of unsupervised methods applied to extract latent information both in text and in user-provided photos. Later, a supervised method was trained to predict review helpfulness based on the text and photos' latent features. Mainly, latent features from the text were extracted through long-short term memory (LSTM) neural networks, which excel at discovering semantic relations between elements of a time-sequence, such as the phrases found in a customer review.

The previous studies provided different alternatives to solve a significant problem: given the vast amount of information found in online hospitality platforms, their users are faced with the impossible task of reading all of the other customers' reviews. Therefore, a way of predicting the helpfulness of customer reviews is wanted to show to the platforms' visitors only the most relevant and helpful comments that will lead that visitor to make a decision and become a customer of the platform. On the other hand, there is also a research branch that is interested in processing the same information but for a different purpose: extracting global features from the customers as an entity to deduce their main concerns and sources of satisfaction. This is the branch that our study belongs to. As might be expected, those general concepts enclosing what worries or satisfies the customers drive the qualitative assessment of the quality of service.

To illustrate,~\citet{CHENG201958} use sophisticated features by exploiting word co-occurrence matrices, which are at the core of the software Leximancer~\citep{Smith2006} that was used for text mining the reviews of Airbnb customers. The outcome of this unsupervised method revealed the central concepts used by customers to review their staying experience. Furthermore, sentiment analysis, which is focused on classifying if the sentiment of an opinion is positive or negative with respect to some topic, was performed on the reviews corpus.

Another exciting application of powerful learning techniques is given in the work of~\citet{LUO2019144}, where latent aspect rating analysis (LARA) was used to identify which aspects related to the staying experience had the larger impact on the rating provided by the customers from a sample of Airbnb reviews. The method is based on the popular Latent Dirichlet Allocation (LDA) model~\citep{Hoffman:2010}, which finds co-occurring relations between words within texts that are simultaneously assigned (probabilistically) to latent topics. These latent topics contain human-level concepts that are automatically extracted from customer reviews. Moreover, the study also performs sentiment analysis to measure how much impact sentiment has on the review score.

All these studies are in line with the robust growth of machine learning applications in the hospitality sector that allows researchers and managers to simultaneously consider thousands of customer reviews to automatically extract the main aspects that concern or satisfy customers at specific locations. The results of these studies are closely related to the measurement of quality of service. In the research reported in this paper, we use some of the methods from these studies to build a data processing pipeline with machine learning state-of-the-art models to target the problem of extracting the main quality of service-related topics from customer reviews. Those models aim to numerically represent the semantics of reviews, which can be used to quantify rating-like levels of customer satisfaction. In what follows, we describe the design of our study as well as the dataset that was retrieved.

\subsection{Data}

Most of the work done in this early stage of machine learning application to customer reviews in the hospitality sector has used data from developed countries, usually including reviews in English solely. Thus, we consider it a vital research step to perform studies with data from non-English speaking countries, not only because of different language structures but also from the cultural perspective, where other sociological aspects from different countries come into play. Therefore, we decided to study two of the main touristic destinations~\citep{citur2020} in a developed and a developing country: we gathered customer reviews from Madrid, Spain and Bogotá, Colombia. The online platform that we chose to gather the data was Booking.com for two reasons. First, it is a popular platform for travellers and tourists to choose and book rooms at hotels. Second, when they gather information from clients at the end of their stay, they ask for positive and negative comments, as well as rating some hospitality aspects. 

Thus, the information that the customer gives to Booking.com in a review is: author's name and country, positive comment, negative comment, cleanliness score, staff score, location score, comfort score, value for money score, facilities score and free WiFi score. The scores range from 1 to 10, 1 corresponding to a bad staying experience and 10 to a pleasant experience. However, each score is not made public by Booking.com at the customer level. It is only available at the hotel level. Therefore, at the customer level, only the average of those scores is available.

We built a web crawler to gather the two cities' data, with a total count of 667 hotels in Bogotá and 1,181 in Madrid. As per the number of reviews, there were 108,563 for Bogotá and 498,361 for Madrid. It is worth noting that the comments are written in several languages, being Spanish and English the most common ones since Colombia and Spain are Spanish speaking countries. We focused on comments written only in Spanish, but the framework we propose can be applied to sets of reviews of any common language~\citep{joshi2020state}.

\subsection{Study design}

From the literature, we encountered a handful of works performing NLP on customer reviews from online hospitality platforms, but none of them focused on the quality of service. Nonetheless, many interesting conclusions can be drawn from these works. One is that deep learning architectures perform much better than classical machine learning algorithms, especially when extracting non-supervised features from the text, i.e. semantic features, instead of engineered features. Another important finding by \citet{LUO2019144} was that a method based on LDA, a topic modelling scheme, helped study aspects and emotions present in customers reviews.

\begin{figure}[!ht]
    \centering
    \includegraphics[width=0.7\textwidth]{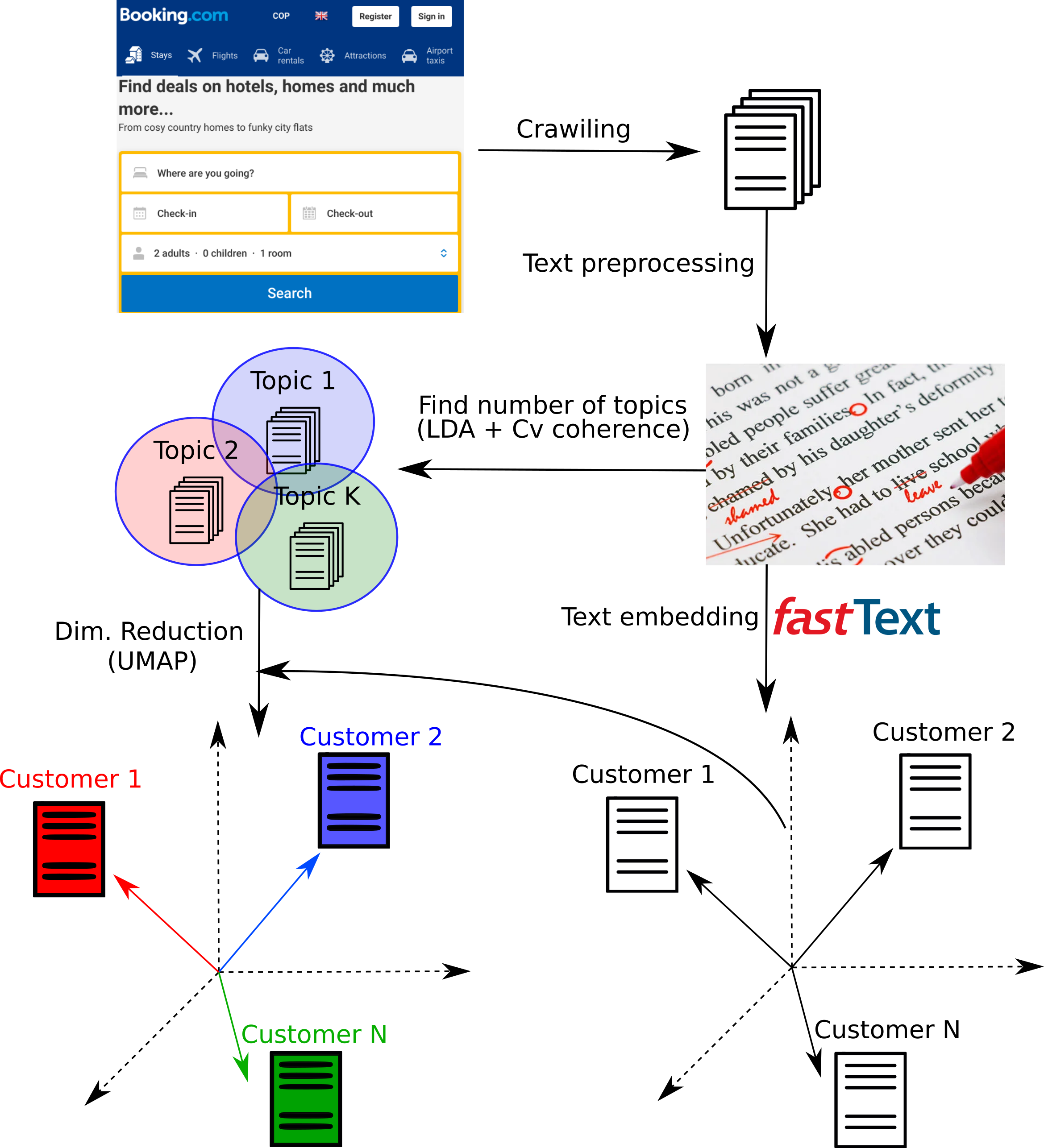}
    \caption{Flow diagram of our work, which facilitates a qualitative assessment of the quality of service. First, customers reviews get crawled from Booking.com. Then a series of text preprocessing steps are carried out on the reviews. After that, two algorithms act on the preprocessed text: one that finds the number of quality of service-related topics in the text corpus, and another one that embeds the reviews in a real vector space. Finally, unsupervised clustering techniques are applied to the embedded vectors to group reviews in different topics at a fine-grained scale.}
    \label{fig:qualflow}
\end{figure}

The previous elements are thus incorporated in the design of our study. We will describe the method that we propose to extract the quality of service-related features from textual information found in clients' comments on Booking.com. In general terms, we discovered the most relevant topics of positive and negative comments posted by customers of all hotels in both cities. We then created a visualisation tool that enabled us to explore the content of the topics with ease. \Cref{fig:qualflow} shows the flow diagram of our study, which focuses on the main NLP task, namely, representing text numerically. We will explain and motivate each step of the study in the following subsections.

\subsubsection{Data extraction and cleaning}

First, we built a crawler using Selenium~\citep{selenium}, a tool for automating web browsing, which enabled us to rapidly download several thousands of customer reviews from all the available listings in Booking.com for Bogotá and Madrid. Second, we performed a text preprocessing, which is intended to clean the text and standardise it. The preprocessing included removing stopwords and punctuation, writing all characters in lowercase and lemmatising each token. As an example, consider the text ``there were always kids playing in the Restaurant!''. Stopwords do not significantly contribute to the semantics of the text, which in the example would be ``there'', ``were'', ``in'' and ``the''. After removing these stopwords, removing punctuation, and writing all characters in lowercase, the remaining words are lemmatised so that the preprocessed text is ``always kid play restaurant''. The lemmatisation was carried out with a Spanish lemmatiser from the open-source software spaCy.

\subsubsection{Topic discovery}

One of the most studied branches in NLP is topic discovery, which aims to identify groups of documents within a corpus that can be distinguished from other documents due to characteristic words that define a topic. Thus, documents that share a topic are semantically similar. For example, a topic can contain customer reviews that complain about the hotel staff kindness, and another topic can contain reviews that praise how quickly the hotel staff meets the customer's requirements.

Defining the number of topics in a corpus is an open research question in machine learning. However, the use of a topic model called Latent Dirichlet Allocation (LDA) combined with the $C_V$ coherence to measure its quality has been satisfactory~\citep{vargasEcuador,vargascaldern2019event} in Spanish opinions corpora. LDA~\citep{Hoffman:2010} is a probabilistic algorithm that groups texts from a corpus into a number $K$ of groups or latent topics. In other words, LDA identifies $K$ topics in a set of texts and assigns probabilities of finding the contents of each topic within each text. To measure the quality or coherence of the topics, i.e. how well-defined a topic is, we use the $C_V$ coherence, a measure of the correlation of top occurrent words in a topic (see~\cref{sec:cvcoh} for details). As an example, if we tell LDA to find five topics in a set of customer reviews, we might find that one topic contains customer reviews saying that the hotel staff never cleans the bathroom, others saying that the towels were of inferior quality. Indeed, those are bathroom-related reviews, but if we had told LDA to find six topics, this topic is likely split into two: one related to the cleanliness of the bathroom and another one related to the quality of the towels. In this case, the $C_V$ coherence is greater than the case where only five topics were used. However, if the number of topics is too high, the $C_V$ coherence decreases because there exists a high similarity between some topics. Indeed, the $C_V$ coherence has shown outstanding agreement with human judgements of the interpretability of the topics extracted by LDA~\citep{Roder:2015,syed2017}. An important remark is that LDA gives the probability that each text belongs to a topic, which allows a text to contain elements from different topics. For instance, a customer may complain about the quality of the bed but can also praise the hotel restaurant chef, which is reflected on the LDA model by assigning to this review a mixed probability of belonging to two topics.

\subsubsection{Visualisation of reviews and topics}

An intuitive and accurate way of exploring an extensive reviews dataset is to navigate through representative reviews of each topic. Here we describe the inner working of a visualisation tool that we designed to explore and read few but representative reviews for every topic that LDA and $C_V$ coherence can discover.

A helpful way of visualising reviews and topics is to consider their vector representations, use a dimensionality reduction algorithm (which takes vectors from a high-dimensional space to a 2-dimensional space), and plot each review as a coloured dot, where the colour indicates the topic. In our case, we use an algorithm called uniform manifold approximation and projection (UMAP)~\citep{2018arXivUMAP}. What UMAP does is to learn topological features of $N$-component vectors to find accurate projections onto a lower-dimensional vector space that tries to preserve the distance from the high-dimensional space to the low-dimensional space. More formally stated, UMAP retains the original vectors' manifold structure in the high-dimensional vector space and projects it onto a low-dimensional vector space.

The resulting plot from reducing the dimension of vectors with UMAP will show a map of the corpus, where points near each other mean that their respective documents are semantically similar. When using the vector representation of LDA, documents from different topics appear somewhat mixed in these plots (not shown) because the vector representation is done in a $K$-dimensional vector space, where $K$ is the number of topics, which in the case of this work is of the order of 10.

Therefore, to improve this, we are motivated to use vector representations of the reviews containing more fine-grained semantic information. An excellent algorithm to do this is the celebrated FastText text vector embedding model~\citep{bojanowski2016enriching,joulin2016bag}, which efficiently uses the co-occurrence of words in the text in order to assign vectors of $N$ components to a piece of text. The vectors store in their components semantic features and linguistic contexts from the text so that two similar texts will have assigned to them two similar vectors. In the lower-left part of~\cref{fig:qualflow}, this is depicted in a 3D vector space (so that $N=3$) with three example customer reviews, but in general, one can choose the dimension $N$ of the vector space where the embedded vectors are. The readers interested in the underlying mechanism of text embeddings are referred to \cref{sec:fasttext} where we give essential concepts of FastText, or to the FastText papers~\citep{bojanowski2016enriching,joulin2016bag} as well as the Word2Vec paper~\citep{mikolov2013distributed}, which give much more mathematical detail about this method. This vector representation achieved by FastText is much richer than LDA's.

Therefore, the reviews and topics' visualisation can be achieved by computing vector representations of the reviews in an $N$-dimensional vector space using FastText ($N=100$ for this work). Then, these vectors are reduced to 2-dimensional vectors using UMAP. Next, a scatter plot is generated, where the colour of each point is determined by how LDA assigns the corresponding review to a topic. 

The only missing detail is that only representative reviews should be plotted, as there might be reviews containing several topics, thus making it challenging to explore specific topics. To filter out only representative reviews for each topic, we choose the reviews that belong to that topic with a probability higher than a selected threshold. In our case, we chose this threshold to be 0.8, although the selection of this value is entirely arbitrary and depends on how many points the user would like to see plotted. It is worth noting that in the topic exploration phase, we ignore those reviews whose maximum probability of being assigned to a topic is less than 0.8; however, the topic probability distribution for each review can be used, for instance, to assess the magnitude of each positive or negative topic within a hotel by computing $M_T = \sum_{r\in R_H}P(r\in T)$, where $M_T$ denotes this magnitude for a topic $T$, $R_H$ is the set of reviews of hotel $H$, and $P(r\in T)$ is the probability that the review $r$ belongs to topic $T$. The analysis of these magnitudes for each topic in a hotel, compared with other hotels, might give a qualitative and quantitative market comparison.

All of this process is depicted in~\cref{fig:qualflow}, and is done in 4 sets of reviews: positive reviews from customers in Bogotá, negative reviews from customers in Bogotá, positive reviews from customers in Madrid and negative reviews from customers in Madrid. Our framework allows us to find topics and visualise reviews in those topics for each of the mentioned groups. Moreover, this framework can be applied to any group of customer reviews: one can consider customer reviews from a whole country, for a single hotel or a group of hotels, etc.

\subsubsection{Scores and topics}

Finally, we will show the score distributions for each topic found in the negative and positive comments in Bogotá and Madrid. Each review is composed of a positive, a negative comment and a score. Therefore, positive topics are not necessarily related to high scores, nor negative topics are necessarily related to low scores. In fact, these score distributions will show us which positive (or negative) topics are essential for the customer to assign a high or low score. For example, a customer might have a terrible experience but might have liked the hotel restaurant's food. Therefore, this customer might give a negative comment, reviewing which aspects he or she did not like about their stay, and also might give a positive comment saying that the food at the restaurant was good; however, since the staying experience was mostly bad, the customer assigns a low score. This means that even though the food was good, it is not a determinant factor to compensate for the other hotel features that the customer perceived as bad. We must emphasise that the perception of a hotel as better, when compared to another, is not driven by standardised service or facilities, but rather it is driven by the average gap between the customers' expectations and perceptions of the quality of service (this work proves this claim).

\section{Results and Discussion}
\label{sec:results}

We will interpret the results provided by our framework by performing constant comparisons to the widely used quality of service model Servqual since it is one of the most researched and robust quality of service models in the hospitality industry. We will see that our framework allows us to recognise some of the Servqual model's dimensions but extends it and limits it to the topics that are more relevant to the customers.

\begin{figure}[!ht]
    \centering
    \includegraphics[width=0.8\textwidth]{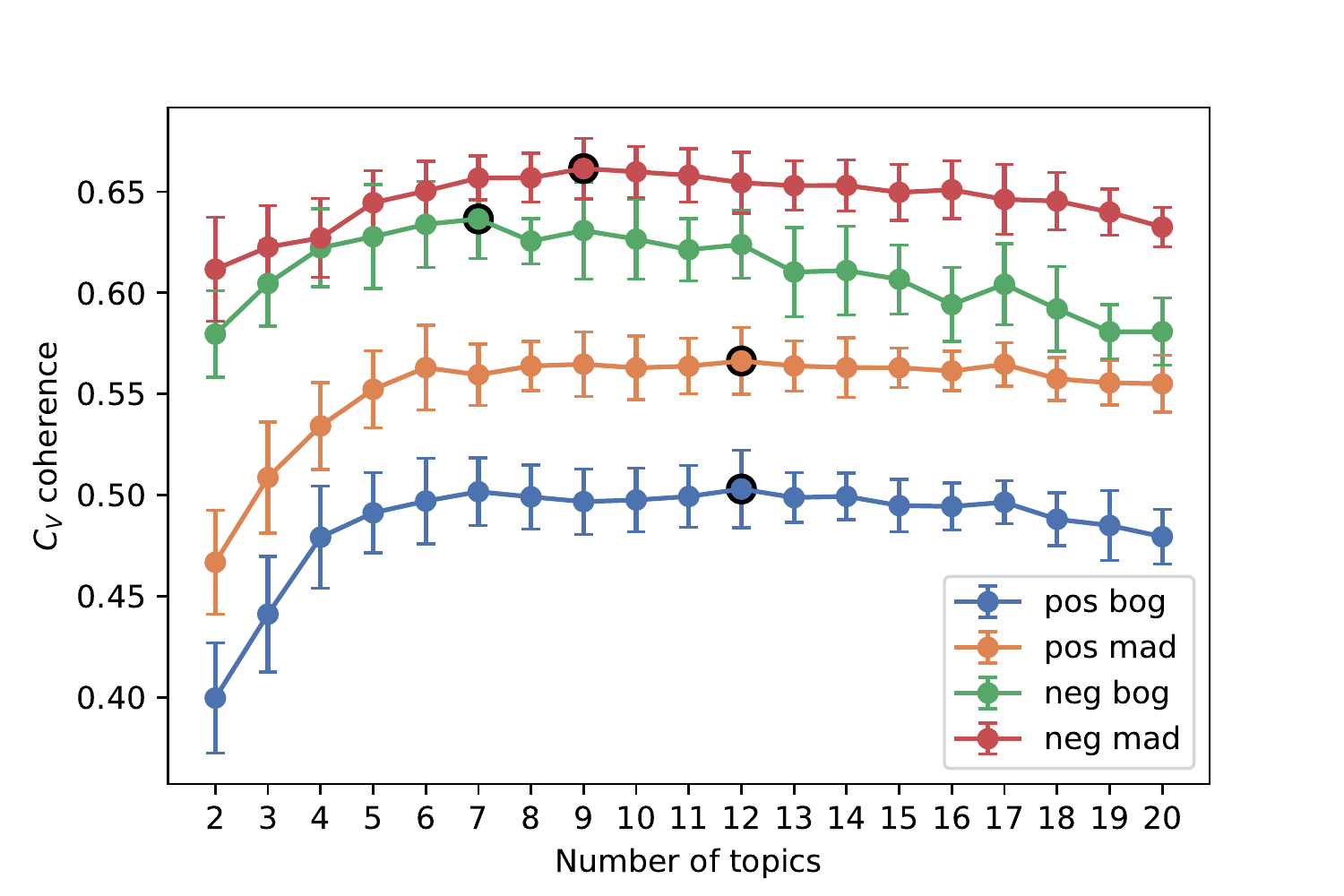}
    \caption{$C_V$ coherence as a function of the number of topics in an LDA model. Since LDA is probabilistic, we measured the coherence by independently training 20 times an LDA model for each number of topics and each set of comments (we found 20 times to be a good number, as adding more training runs did not affect coherence variance). The data points are the average of those measurements, and the error bars are their standard deviation. Circled in black are the maximum coherence for each set of reviews.}
    \label{fig:coherences}
\end{figure}

The first step of our method aims to estimate the number of topics in each of four customer reviews sets: positive and reviews from Bogotá and Madrid. \Cref{fig:coherences} shows the measurement of the $C_V$ coherence as a function of the number of topics, where the coherence reaches a maximum for 12, 12, 7 and 9 topics for positive reviews from Bogotá, positive reviews from Madrid, negative reviews from Bogotá and negative reviews for Madrid, respectively.

\begin{figure}[!ht]
    \centering
    \subfloat{\includegraphics[width=0.88\textwidth, trim={2cm 4.5cm 2.1cm 7.5cm},clip]{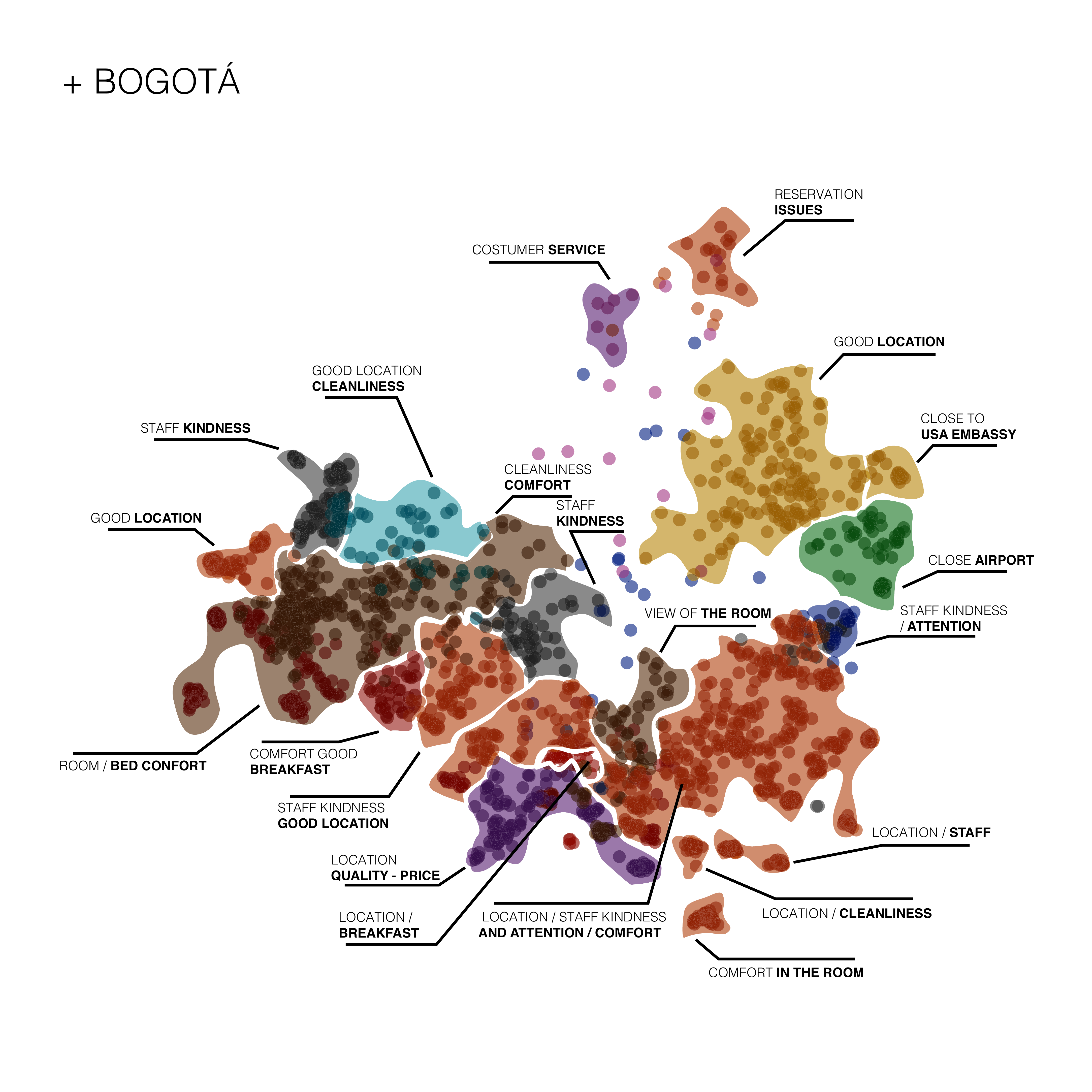}}\\%
    \subfloat{\includegraphics[width=0.88\textwidth, trim={4cm 5.2cm 2.2cm 9cm},clip]{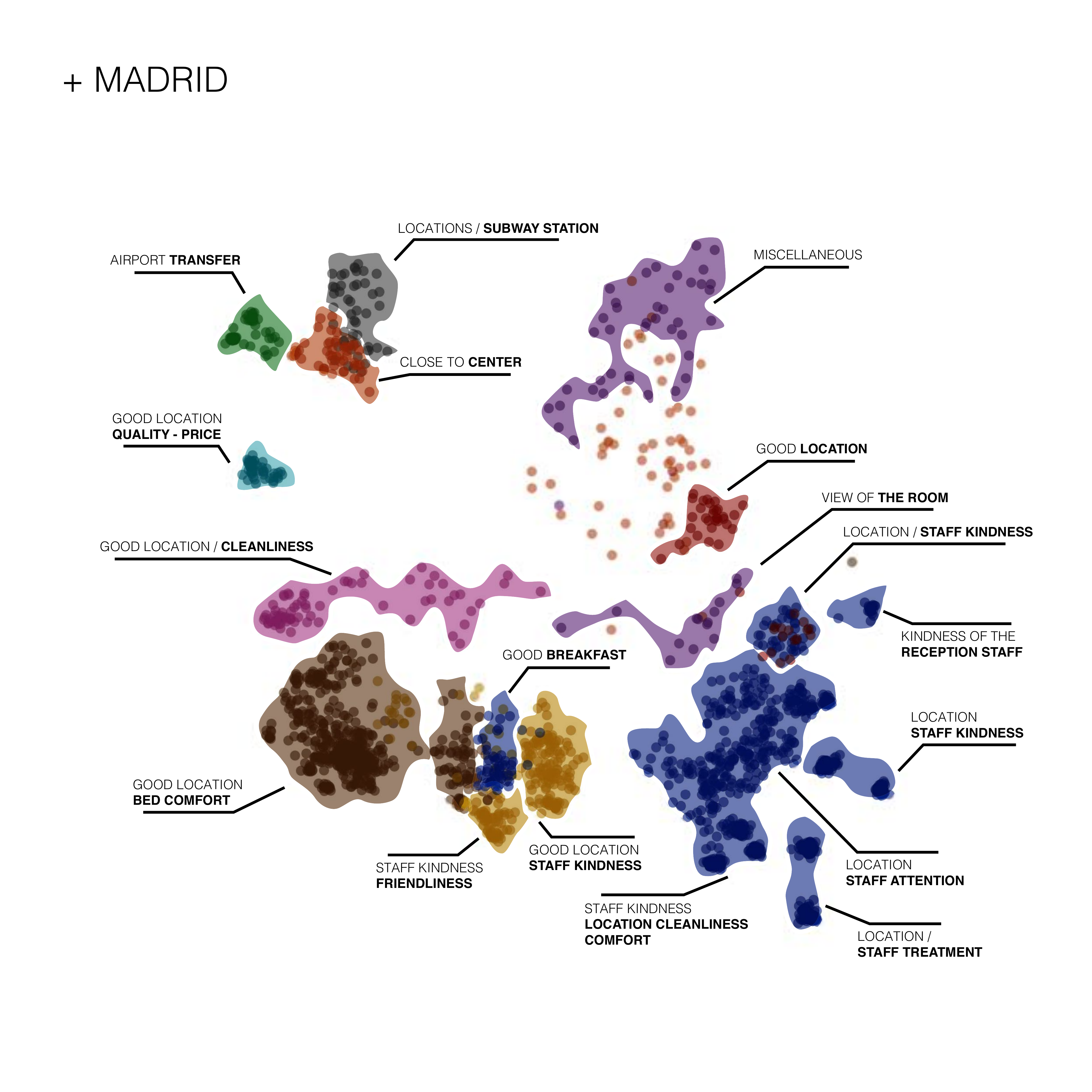} }%
    \caption{2D projection found by UMAP of clusters coloured by LDA analysis of the FastText representation of positive reviews from Bogotá (top panel) and Madrid (bottom panel). Each point in the figure represents a single customer review.\label{fig:pos_visualisations}}%
\end{figure}

After obtaining the FastText vector for each review and performing dimensionality reduction with UMAP, we ended up with visualisations that allowed the exploration of representative reviews for each of the identified topics. The visualisations of positive topics from Bogotá and Madrid are shown in \cref{fig:pos_visualisations}, where each point in the figures represents a single customer review, representative of a latent topic. For Bogotá, 9.5\% of the positive reviews were representative (i.e. the highest probability to be assigned to a topic surpassed 80\%), and for Madrid, 6.4\% of the positive reviews were representative. The colour of the points identifies different latent topics. Points close to each other have a similar FastText vector representation, which is why each figure contains several well-packed clusters, separated from the rest of the comments. This visual representation allows a better human interpretation of the topics that are important for hotel clients. The clusters are easily identifiable and have well-defined topics, which were annotated and shown in the figures.

We see that there are many similarities in the topics found in positive reviews from Bogotá (top panel of \cref{fig:pos_visualisations}) and Madrid (bottom panel of \cref{fig:pos_visualisations}), but we also see some differences. In general, reviews provide information that is not usually considered by quality of service models in the hospitality industry, such as adaptations of the widely used Servqual model. Nonetheless, we emphasise that all of this information is contained within the general dimensions defined by the Servqual model. Most of the positive reviews refer to variables included in the tangible elements Servqual dimension, meaning that having comfortable bedrooms, clean spaces, beautiful room views and good breakfasts are crucial for the clients when giving a positive review. Also, the kindness of the staff is vital in positive reviews for both cities. However, there are other elements of great value for the clients which are not usually explicitly included in the Servqual model, such as good location, airport transfer services, closeness to metro stations, closeness to the centre (in Madrid), closeness to USA Embassy or airport (in Bogotá). These topics, however, can also be associated to the tangible elements Servqual dimension. Also, the booking system seems to be meaningful and valued in Bogotá, and we suppose that clients in Madrid assume an excellent booking system as given.

\begin{figure}[!ht]
    \centering
    \subfloat{\includegraphics[width=0.83\textwidth, trim={6.1cm 6.5cm 2.5cm 7.8cm},clip]{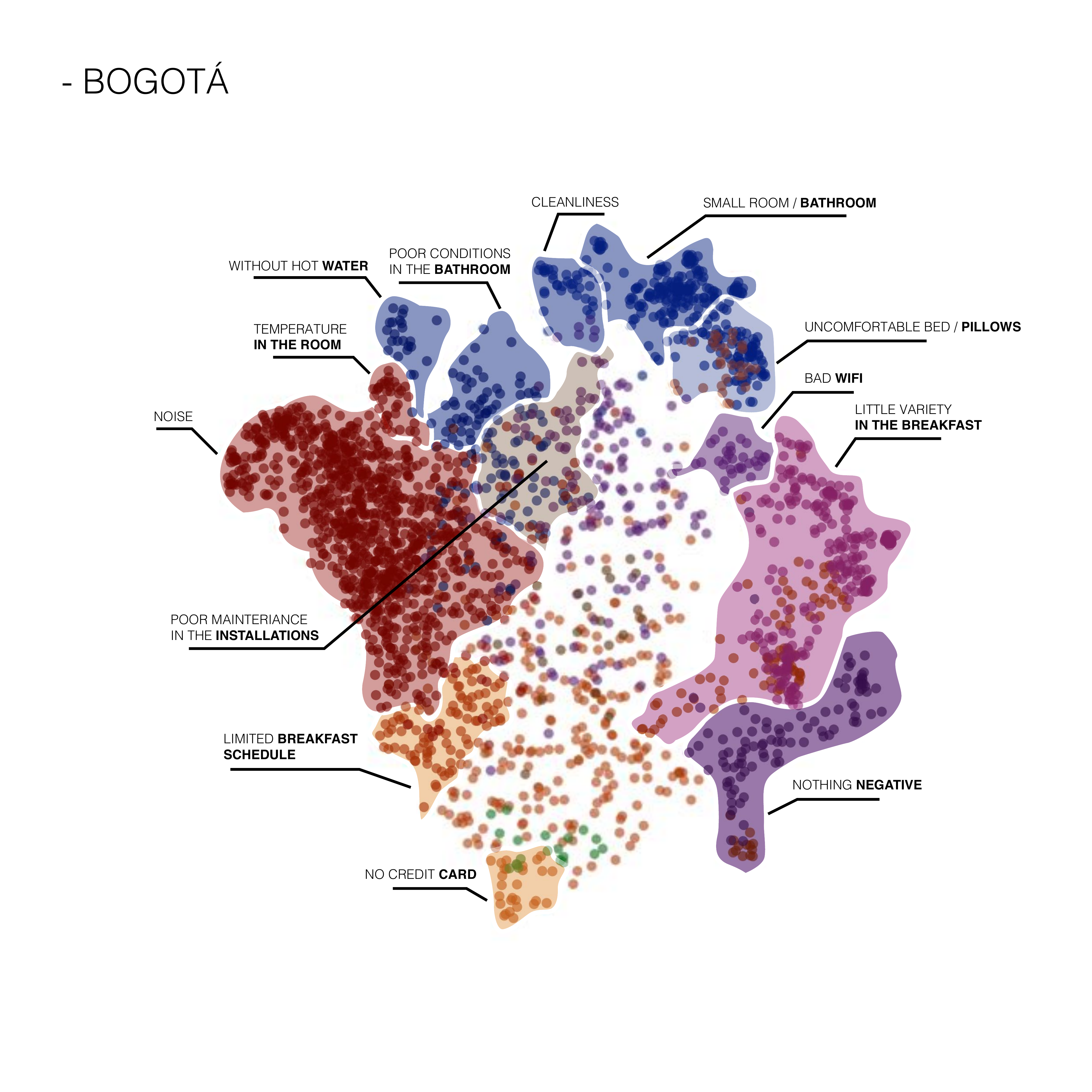}}\\%
    \subfloat{\includegraphics[width=0.83\textwidth,trim={2.7cm 3cm 2.1cm 6cm},clip]{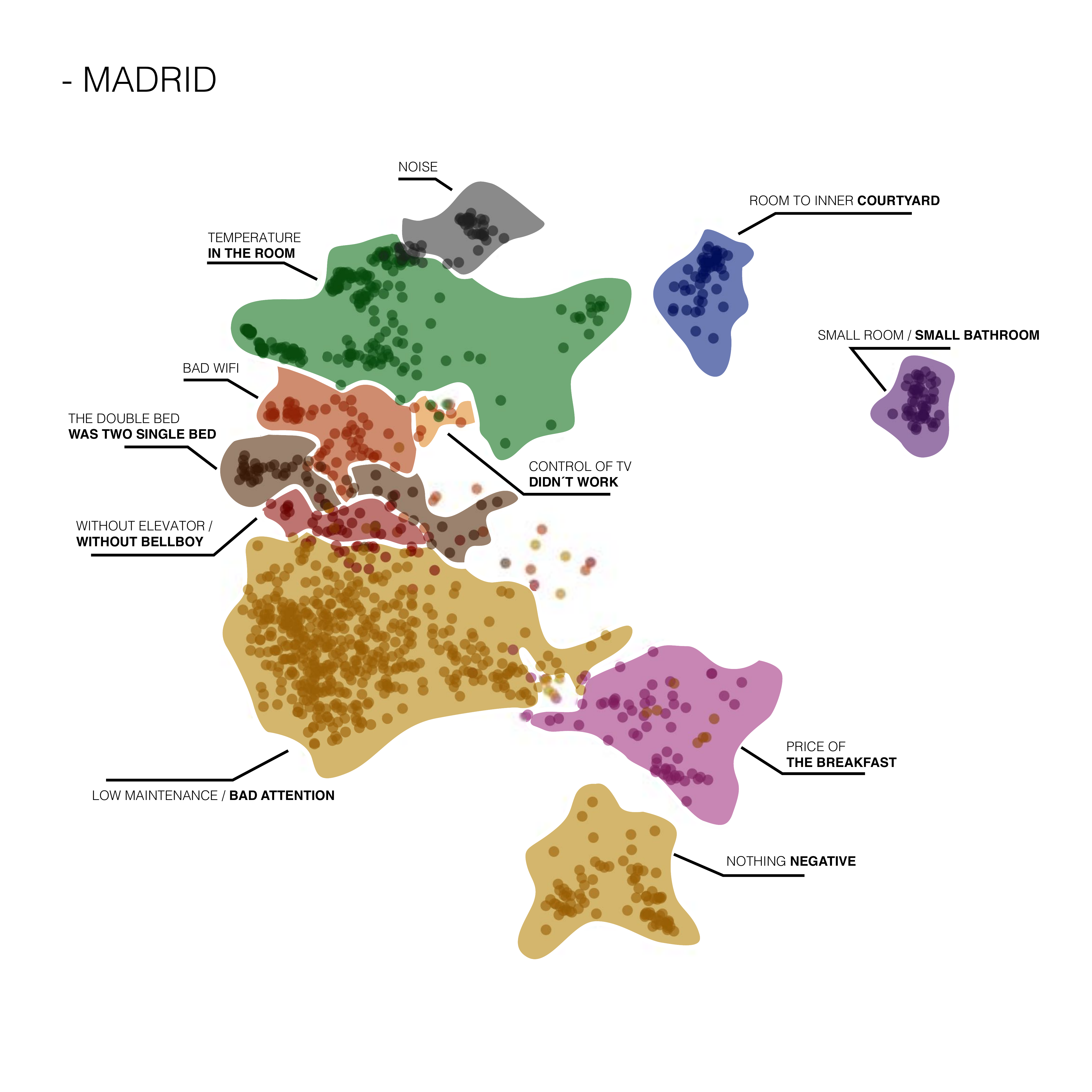} }%
    \caption{Same as \cref{fig:pos_visualisations} but for negative reviews from Bogotá (top panel) and Madrid (bottom panel).\label{fig:neg_visualisations}}%
\end{figure}

These examples show that our method automatically shows the main topics on which hotels should focus their attention to induce in their clients a positive perception of their stay and can do so at any granularity level (in this case, at a city level). This is further taken advantage of when examining the negative reviews, as shown in \cref{fig:neg_visualisations}, where 14.3\% of the negative reviews were representative for Bogotá 7.5\% for Madrid.

In the case of negative reviews, most topics also refer to Servqual's tangible elements dimension. Some of these topics are temperature in the room, poor maintenance in the installations, small room/bathroom, noise, and lousy WiFi. In Bogotá (top panel of \cref{fig:neg_visualisations}), clients complain about uncomfortable bed/pillows, no credit card payment option, little variety in breakfast or a limited breakfast schedule. 

On the other hand, in Madrid (bottom panel of \cref{fig:neg_visualisations}), topics such as room facing the inner courtyard or lack of an elevator and lack of bellboy stand out. It is observed that negative reviews stress cultural differences. Bogotá contains many negative reviews related to noise, which is a factor that must be taken care of by Bogotá's hospitality sector by sound-proofing hotel rooms. As the food is relatively cheap in Colombia, no price complaints are made in Bogotá, but they highlight in Madrid. Also, the option of paying with a credit card is not a central topic in Madrid, but it is in Bogotá. This can be due to a cultural aspect of Colombia, where many local hotels do not support credit card payment\footnote{As a matter of fact, in the past two years, Colombian government is applying a public policy where all kind of businesses have to have electronic payment options for electronic billing (see Ruling 42 of 2020 from Colombia's DIAN).}.

To help the reading of \cref{fig:pos_visualisations,fig:neg_visualisations}, \cref{tab:posneg} lists the main topics found in the positive and negative reviews of both cities, along with their associated Servqual dimension. The association of Servqual dimensions to each major topic found through our framework is done by examining the relation between relevant comments for each topic and the definition of each Servqual dimension. As explained by \citet{zeithaml2018services}, there are five Servqual dimensions:
\begin{itemize}
    \item \textit{Tangible elements}. These are related to physical appearance, physical facilities, infrastructure, equipment, materials and staff.
    \item \textit{Reliability}. It refers to fulfilling the promised service to the client carefully and reliably. This implies that the company is faithful to every feature of the service they say they will provide.
    \item \textit{Responsiveness}. This refers to being available to help clients and to provide adequate and quick assistance in any need (question, complaint, problem) related to the service being provided.
    \item \textit{Assurance}. This is the capacity to inspire credibility and trust in the client based on the employee's knowledge, skill and attention.
    \item \textit{Empathy}. It refers to the custom attention offered by the company to its clients. It must be transmitted through a custom service, adapted to the clients' needs.
\end{itemize}

These definitions allow the categorisation of fine-grained topics found in~\cref{fig:pos_visualisations,fig:neg_visualisations} into the Servqual dimensions. However, the fine-grained topics are only aspects of more prominent topics, which were automatically identified with our framework, and shown in~\cref{tab:posneg}. In what follows, we will provide an association between major topics and Servqual's dimensions.

For positive reviews, we see that the following topics are related to tangible elements: bed comfort, large bedroom, cleanliness, good breakfast, and view of the room. The airport transfer-related topic can be classified into the empathy Servqual dimension if the hotel presents a service to their customers to ease airport transfers; but, if the hotel possesses buses or cars to provide such a service, this topic can also be related to the dimension of the tangible elements. Reservation issues and their solutions, as well as the miscellaneous topic (this topic includes aspects related to problems about facilities that were effectively solved by the hotel staff), are related to the responsiveness Servqual dimension. Staff kindness or attention and quality-price relation are related to reliability. Customer assistance is related to empathy. Location-related topics is a significant finding of our study: customers value how near the hotel is from certain points of interest within the city or how easy it is to get from the hotel to those points of interest. We should mention that, even though Booking.com asks customers to evaluate the hotel location, our framework identifies this topic without \textit{a priori} knowledge about this request. These aspects are not usually related in other studies to tangible elements, but we identify that they can be included in this Servqual dimension. However, it is important to remark that tangible elements use to be linked to features inside the hotel, whereas location is related to the hotel position within the city relative to points of interest. Thus, location-related topics become also an important feature for the hotel's value proposal, especially when promoting the hotel, for instance, through the Booking.com website. Moreover, some location topics that are related to security perception (in the criminal sense), could also be associated to the assurance Servqual dimension.

Regarding negative reviews, room and bathroom issues are related to tangible elements and reliability, as sometimes hotels do not provide the same offered services. Even though some of these issues can be directly associated with the tangible elements Servqual's dimension, there are specific aspects captured by our framework that do not tend to be in customer service surveys, such as the availability of hot water or the bathroom size. Also, in the topic of room and bathroom issues, there is quite a common reliability-related issue that stands out in the aspects retrieved by our framework, which is that hotels mistakenly provide two individual beds when one double bed was booked. Noise (absence of sound-proof environments), lousy WiFi, no elevator, poor maintenance in the installations, payment methods (e.g. the unavailability of credit card payments),  and breakfast issues can also be classified in the tangible elements dimension. Billing issues are related to two Servqual categories: reliability and assurance. Commonly, clients express billing issues because there were errors in the billing, which increase distrust in the hotel staff. Regarding empathy, it encloses topics such as bad attention. An interesting topic is dangerous location, which is usually not related directly to any variable in the several adaptations of the Servqual model to the hospitality industry. Reviews related to this topic show that the hotel is located in places perceived as insecure. Therefore, the dangerous location topic is more related to the hotel's environment, falling into the tangible elements and also assurance Servqual's dimensions. Hotel management should be aware of this situation to improve the safety perception of their clients.

Moreover, \cref{tab:posneg} presents the percentage of reviews related to each topic, as well as the salience of each topic, which we compute by summing the probabilities that a word belongs to a topic considering the top 10 words for each topic. This table shows one of our main results: there are important topics for the customers that are usually not found in the surveys from different adaptations of the Servqual model, such as the location, with a significant percentage of positive reviews: 34.8\% for Bogotá and 21.8\% for Madrid. Also, \cref{tab:posneg,fig:pos_visualisations,fig:neg_visualisations} show that some service elements are not equally valued in different countries because distinct tourist populations have diverse expectations. An example of this is the clear presence of topics related to billing issues or payment methods in Bogotá and not in Madrid. Another notable point is the large amount of concern displayed in the reviews regarding room/bathroom issues such as temperature in the room (in Madrid), no hot water (in Bogotá), room/bathroom size, poor bathroom conditions, wrong bed size, among others. Indeed, this is a neuralgic point, especially in Madrid, taking almost half of the negative reviews.

\begin{table}[!ht]
    \centering
    \caption{Main latent topics depicted in  \cref{fig:pos_visualisations,fig:neg_visualisations} for positive and negative reviews and their associated Servqual dimensions. The first column presents the main major topics for positive and negative reviews. The second column relates Servqual dimensions to those topics. Rows are grouped with green/white rows based on those Servqual dimensions. The last two columns present numbers of the form $XX/YY$, where $XX$ stands for the probability percentage assigned to a topic by the LDA model taking into account all documents, and $YY$ stands for the salience of each topic, by summing the probability of the top 10 words assigned to each topic by the LDA model.
    \label{tab:posneg}}
    \begin{tabularx}{\textwidth}{bmss}
    \toprule
         & & \multicolumn{2}{c}{City}  \\
         \cmidrule(r){3-4}
         \textbf{Positive reviews}& Servqual dimension & \heading{Bogotá} & \heading{Madrid}\\
         \midrule
         \rowcolor{green!15}
         Bed comfort & & 12.6/0.63& 13.8/0.69\\
         \rowcolor{green!15}
         Large bedroom & & \cellcolor{gray}& 7.3/0.24\\
         \rowcolor{green!15}
         Cleanliness & & 8.6/0.43&8.1/0.47\\
         \rowcolor{green!15}
         Good breakfast & & 4.3/0.41&6.8/0.32\\
         \rowcolor{green!15}
         Location-related & & 34.8/0.47  & 21.8/0.35\\
         \rowcolor{green!15}
         View of the room & \multirow{-6}{*}{Tangible elements} & 8.5/0.45 & \cellcolor{gray} \\
         Airport Transfer & Tangible elements or Empathy & \cellcolor{gray} & 7.3/0.22\\
         \rowcolor{green!15}
         Reservation issues &   &   5.5/0.30&\cellcolor{gray}\\
         \rowcolor{green!15}
         Miscellaneous & \multirow{-2}{*}{Responsiveness}& \cellcolor{gray}&4.9/0.27\\
         Quality-price  &   &   6.1/0.48&5.0/0.71\\
         Staff kindness/attention &   \multirow{-2}{*}{Reliability} & 12.9/0.42 &25.0/0.71\\
         \rowcolor{green!15}
         Customer assistance & Empathy &   6.9/0.24&\cellcolor{gray} \\
         \midrule
         \textbf{Negative reviews }&&&\\
         \midrule
         \rowcolor{green!15}
         Room/bathroom issues & Tangible elements and reliability& 19.8/0.30&44.9/0.30\\
         Noise & & 20.1/0.23&9.8/0.37\\
         Bad WiFi & & \cellcolor{gray}&9.0/0.18\\
         No Elevator & & \cellcolor{gray} &7.9/0.32\\
         Poor maintenance in the installations &  & 11.0/0.18&\cellcolor{gray}\\
         Payment methods & & 11.3/0.15&\cellcolor{gray}\\
         Breakfast issues & \multirow{-6}{*}{Tangible elements}& 14.2/0.33&14.4/0.20\\
         \rowcolor{green!15}
         Billing issues & Reliability and assurance & 15.2/0.18 &\cellcolor{gray}\\
         Bad attention & Empathy & \cellcolor{gray} & 14.0/0.17\\
         \rowcolor{green!15}
         Dangerous location & Tangible elements and assurance & 8.4/0.16 & \cellcolor{gray}\\
         \bottomrule
    \end{tabularx}
\end{table}

On the other hand, from a linguistic point of view, each topic's salience expresses how unique the vocabulary used to describe positive or negative issues is. For instance, very high saliences are obtained for bed comfort and quality-price relationship in positive reviews concerning other topics. Regarding negative reviews, the largest saliences are related to room/bathroom and breakfast issues and noise (in Madrid).

\begin{figure}[!ht]
    \centering
    \includegraphics[width=\textwidth]{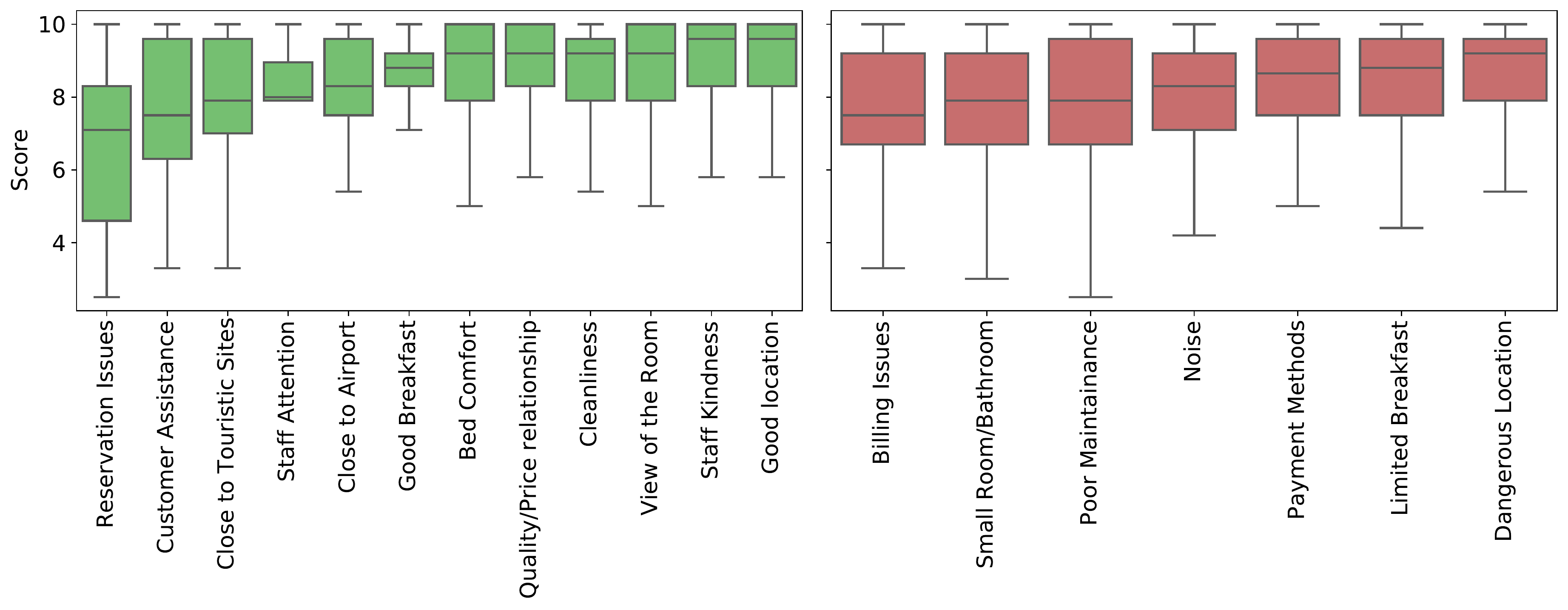}
    \caption{Score distribution of relevant reviews with positive (left) and negative (right) topics for hotels in Bogotá. As before, relevant reviews are selected as those whose probability of belonging to a topic is above a threshold (80\%).}
    \label{fig:score_dist_bog}
\end{figure}

Finally, we study how these identified topics are related to the scores given by the customers. Notably, we answer how frequently are positive and negative topics related to different ranges of scores. To do this, in~\cref{fig:score_dist_bog} we show the score distribution (through box-plots) of the positive and negative relevant reviews identified for each positive or negative topic in Bogotá. In both types of reviews, we found statistically significant differences between the scores using the ANOVA test ($p \ll 0.05$). Furthermore, we performed the Tukey's honestly significantly differenced (HSD) test~\citep{tukey1949comparing} on the different pairs of topics for positive and negative reviews, finding almost all of the differences statistically significant ($p < 0.05$). We see that a group of positive reviews (recall that customers are asked to write a positive and a negative comment) related to reservation issues and how they were solved have a low score distribution with respect to other positive topics. Also, topics related to staff (customer service, staff attention and staff kindness) offer a fascinating insight: while customer service and staff attention get mentioned in positive reviews, their score distribution is much lower than the staff kindness topic score distribution. We hypothesise that customers expect some standard customer service and staff attention, but they really value kindness, which is why this topic is correlated with very high scores. Regarding negative reviews, there is a small topic (meaning that only 8.4\% of reviews consider this topic, not highlighted in \cref{fig:neg_visualisations}) with reviews about the surroundings of the hotel being dangerous (in a security and criminal sense) that has a high score distribution. This means that people do not penalise with low scores when the hotel is situated at a dangerous-looking location but give more importance to other positive aspects of the hotel. On the other hand, topics with a larger percentage of relevant reviews, such as having small rooms or bathrooms, billing issues and noisy rooms, obtain lower scores. In Bogotá, these topics are the most common concerns among customers and are correlated with low scores (below 6/10).

Regarding Madrid, differences in scores through ANOVA and Tukey's HSD test were also statistically significant $(p \ll 0.05)$ between topics. As in Bogotá, good location and staff kindness/attention are important positive topics correlated with high score distributions. Interestingly, many customers point out that rooms are spacious, but this is a feature that is not necessarily correlated with high scores. With respect to negative reviews, small bedrooms or bathrooms, wrong bed sizes (e.g. customers book a double bed but get two single beds instead), and bad attention from the staff are main topics of concern for customers correlated with low score probabilities. On the other hand, sometimes hotels do not include breakfast by default, and many customers find this disappointing. Thus, they highlight this issue as a negative topic when they give low scores (below 6/10). Also, there is a high frequency of negative reviews associated with street noise for low scores, meaning that it is an essential source of discomfort for customers.

\begin{figure}[!ht]
    \centering
    \includegraphics[width=\textwidth]{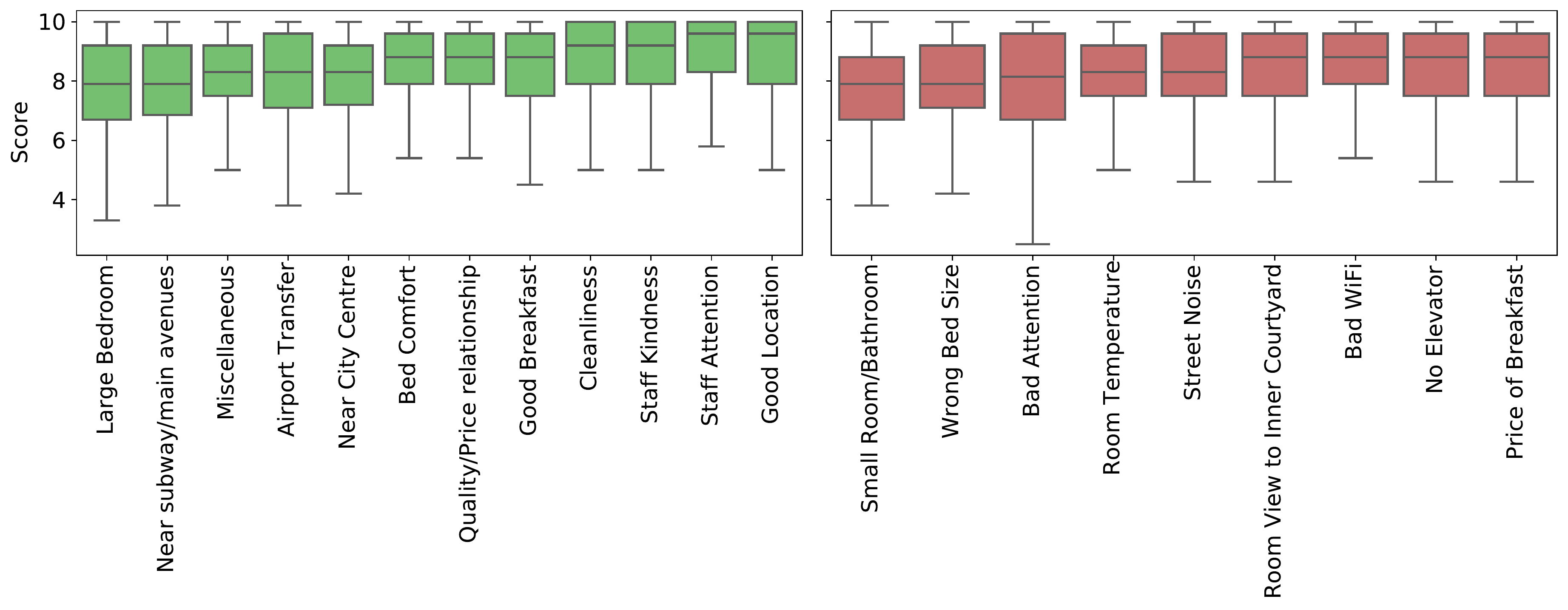}
    \caption{The same as~\cref{fig:score_dist_bog} but for Madrid.}
    \label{fig:score_dist_mad}
\end{figure}

\section{Conclusions}
\label{sec:conclusions}

In this work, we presented a framework to automatically extract the quality of service-related features from large databases of hotel customer reviews. We exemplified the use of our framework with customer reviews from Bogotá, Colombia and Madrid, Spain, in Booking.com. The framework is based on machine learning algorithms for latent topic discovery, dimensionality reduction and vector representation of text. By combining these algorithms, we generated visualisations of big data sets of comments by clustering them and showing the most representative ones in each cluster. The results obtained show that the visualisations aid human readability of many comments and show essential information, which could aid decision-making in the hospitality sector. Moreover, we were able to discover the critical aspects for clients when reviewing positively or negatively hotels in the studied cities.

\subsection{Research implications}
Our research's theoretical implications aim to discuss the validity of using large online datasets from online hospitality platforms when assessing the quality of service through non-structured data. Quality of service models have been characterised by a static set of aspects based on customers experiences that are measured and gathered by researchers. The availability of large datasets in online hospitality platforms is changing and will change the methodologies for collecting information and for designing quality of service models. Moreover, the power of machine learning algorithms is starting to be used as a means of building quality of service models that are easily adaptable to the dynamics of the hospitality sector. This work contributes to this new but promising area of research.

Also, in our analyses, we encountered that cultural differences in segments of the global population produced different expectations of their hotel stay. An interesting observation is that Colombian clients rated (from 1 to 10 in Booking.com) Madrid hotels about 5\% higher than the average Spanish hotel client, whereas Spanish clients rated Bogotá hotels about 5\% lower than the average Colombian hotel client. This can also be seen in the comments, as there are different main topics in the two cities. For instance, many people complain about noise in Bogotá (probably because not all hotels have soundproof rooms), and in Madrid, they do not as much. Another example is that clients value being close to metro stations in Madrid because it is an efficient and comfortable public transportation system. On the other hand, in Bogotá, the public transportation system is not efficient nor comfortable and does not densely connect many parts of the city. 

Our framework also has tremendous practical implications. It directly enables managers and researchers to identify the critical quality of service topics that affect (positively or negatively) the perceived quality of service, thus providing essential information to propose and implement improvement strategies. In fact, we showed that some of these topics are not usually taken into account by well-established quality of service models such as adaptations of the Servqual model to the hospitality industry.

Moreover, the topics discovered and explored through our framework can be compared with customer ratings. Although ratings are not as telling and expressive as customer reviews, they give a rough estimate of the customers' level of satisfaction. Therefore, from the automatically identified topics, one can establish a connection between some topics and rating distributions, as was also shown in this work.

\subsection{Limitations and future work}


Although our research used a large dataset, it only explored two capital cities, one online hospitality platform and one language: Spanish. Our framework has the capability to escalate to many other destinations and other platforms such as Airbnb and other P2P and business-to-consumer (B2C) platforms. It would be important to analyse the relevance of including other languages in future research to identify not only cultural differences but also social, psychological and personal differences that influence the perception of service quality of hotel customers in different cities.

Since the amount of text available from customer reviews influences the robustness of machine learning methods, it is important to note that our method could be limited to scenarios in which hotels do not have enough customer reviews. This aspect will be addressed in future work to analyse the effectiveness of our method in small/not popular/low-class hotels with few reviews with respect to high/popular/famous/high-class hotels with many reviews. In this scenario, we remark that doing a stratified analysis by hotel category might reveal topics that are important for certain clientele, but not for others.

Future work includes gathering customer reviews from other cities to achieve better latent topic models. More importantly, we aim for a pilot implementation of this framework as a quality of service assessment tool in a hotel willing to execute plans to improve the quality of service perception by analysing the information provided by our framework.

Finally, a crucial research avenue is a theoretical and practical study of integrating large online hospitality platforms as information sources for robust quality of service models such as Servqual. We foresee two main research lines: how to automatise retrieval information to score the variables of the Servqual model from online reviews; and how to adapt the Servqual model to specific locations and sub-markets, where customers worry about some specific service features.

%
\section*{Conflict of interest}
The authors declare that they have no conflict of interest.


\bibliographystyle{spbasic}      
\bibliography{sample}   

\clearpage
\appendix

\section{Essentials of FastText}\label{sec:fasttext}

FastText is a library that creates text embeddings. This means that a string $s$ is mapped to a vector in the vector space $\mathbb{R}^N$. The FastText method shares the embedding ideas from other models such as Word2Vec~\citep{mikolov2013distributed}. In what follows, we will see the general ideas of how the map/embedding is built; however, the interested reader is referred to~\citep{bojanowski2016enriching,joulin2016bag,mikolov2013distributed,vargascaldern2019event} for a more formal exhibition of the method with mathematical details.

Consider a dataset of texts, or documents $\mathcal{D}$. We can create the vocabulary set $\mathcal{V}$ as the set of words contained in the documents. We can order this set arbitrarily, but for the sake of simplicity, let us assume that we deal with a vocabulary that is alphabetically ordered. Let $V=|\mathcal{V}|$ be the size of the vocabulary. Consider a one-hot encoding map $\phi: \mathcal{V}\to\mathbb{R}^V$ be defined as a function that takes the $i$-th element of the vocabulary (in alphabetical order) and maps it to a vector $\boldsymbol{\phi}_i$, which has all of its components equal to 0 except the $i$-th component, which is equal to 1. The embedding is an $N\times V$ matrix $W$ that maps a vector from the one-hot encoding vocabulary in $\mathbb{R}^V$ to the embedded vector space $\mathbb{R}^N$, where $N \ll V$. This means that the $i$-th word of the vocabulary will have an embedded vector representation $\boldsymbol{w}_i := W\boldsymbol{\phi}_i$ (note that $\boldsymbol{w}_i$ is just the $i$-th column of $W$). The main feature is that words that are semantically similar, also have similar vector representations in the embedded space, i.e. $\boldsymbol{w}_i\cdot \boldsymbol{w}_j /(||\boldsymbol{w}_i|| \, ||\boldsymbol{w}_j||) \approx 1$ for similar words $w_i,w_j\in\mathcal{V}$.

The question that immediately arises is: how can one measure semantic similarity?~\citet{mikolov2013distributed} define semantic similarity with a prediction problem that has its origin in the distributional hypothesis of linguistics~\citep{harris1954distributional}, which states that semantically similar words are used in similar contexts. For instance, the words ``kindness'' and ``courtesy'' are expected to have similar vector representations because they can be found in positive comments about hotel staff with similar contexts. The context is formally defined as the set of words that surround the word of interest, and the amount of words that are taken into the context is normally referred as the context size. The definition of context allows us to state the prediction problem that defines the semantic similarity: given a context around a word of interest $w_i$, can we predict that the word of interest is $w_i$? or, given a word of interest $w_i$, can we predict its context? These two questions are answered by the continuous bag of words (CBOW) and the skip-gram configurations of Word2Vec-like architectures, respectively.

As an example, let us consider the CBOW configuration. Consider a part of a sentence consisting of a word of interest $w$ (we drop the sub-index) and a context of size 4: $w_1\,w_2\,w\,w_3\,w_4$. In the CBOW configuration, we use the context words to predict the word of interest. This is done by averaging the vector representation of the context words, i.e. $\boldsymbol{w}_c = \frac{1}{4}\sum_{i=1}^4 \boldsymbol{w}_i$. The prediction of the word of interest\footnote{Here the prediction is made with the same weight matrix $W$. However, in practice, the prediction matrix has different weights, meaning that there are two different vector representations for each word.} is done by computing $W^T\boldsymbol{w}_c$, which should equal to the one-hot encoding of the word $w$. The matrix elements of $W$ can be learnt through any minimisation algorithm of a loss function such as categorical cross-entropy, built by sampling pairs (word of interest, context words) and predicting words of interest given their context words.

FastText~\citep{bojanowski2016enriching} leverages this idea to learn sub-word information embeddings. Instead of dealing with a vocabulary of words, FastText considers a vocabulary of $n$-char chains. To understand this, consider a sentence which contains the word ``kindness''. We use two special characters $\langle$~and~$\rangle$ to mark where a word starts or ends, so that ``kindness'' is transformed to ``$\langle$kindness$\rangle$''. If we consider $5$-char chains, we would get the following decomposition of ``kindness'': $\{$``$\langle$kind'', ``kindn'', ``indne'', ``ndnes'', ``dness'', ``ness$\rangle$''$\}$. We learn a vector representation for each $5$-char chain found in our vocabulary in the same fashion of context words, and now, the representation of a word is the average of the representation of the chains that form its decomposition. This can be extended to sentence representation by also averaging its word representations.

\section{$C_V$ topic coherence}
\label{sec:cvcoh}

The $C_V$ topic coherence~\citep{Roder:2015} is a metric that correlates well with human topic ranking, which gives a gold standard of interpretability. The $C_V$ coherence is calculated as follows. For each topic, consider the set $W=\{w_1,\ldots, w_N\}$ of the $N$ most frequent words within the documents assigned to that topic. Compute $p(w_i)$ as a frequency that tells the probability of finding word $w_i$ in te documents of that topic. Also, compute $p(w_i, w_j)$ of finding $w_i$ and $w_j$ within a document, with the constrain that $w_j$ must be at most $s$ tokens away from $w_i$, where $s$ is some fixed window size. 

Now, we consider a segmentation of $W$, in the sense of~\citet{douven2007measuring}. Such a segmentation is a set of pairs of subsets of $W$. In particular, the $C_V$ coherence uses a segmentation of the form
\begin{align}
    S = \{(W_\beta', W) \,\vert\, W_\beta \in \mathscr{P}(W)- \{\varnothing \}\},
\end{align}
where $\mathscr{P}(W)$ is the power set of $W$. We refer to each pair in $S$ by $S_\beta = (W_\beta', W)$.

We can represent each set of vectors $\bar{W}\in \mathscr{P}(W)- \{\varnothing \}\}$ with a context vector $\boldsymbol{v}(\bar{W})$ of size $|W|$, whose $j$-th component is
\begin{align}
    v_j(\bar{W}) = \sum_{w_i\in \bar{W}} \text{NPMI}(w_i, w_j)^\gamma
\end{align},
where where NPMI stands for normalised point-wise mutual information and $\gamma$ assigns greater values to larger NPMI's. The NPMI is defined via
\begin{align}
    \text{NPMI}(w_i,w_j)^\gamma = \left(-\frac{\log\frac{p(w_i,w_j) + \epsilon}{p(w_i)p(w_j)}}{\log(p(w_i,w_j) + \epsilon)}\right)^\gamma,\label{eq:nmpi}
\end{align}
where $\epsilon$ is a parameter added for numerical stability. Notice that the numerator in~\cref{eq:nmpi} is just (ignoring $\epsilon$) $p(w_i|w_j)/p(w_i)$, which will be greater than 0 if the conditional probability of $w_i$ given $w_j$ is greater than the probability of the word $w_i$. Therefore context vectors represent the level of co-ocurrence of a set of words $\bar{W}$, with respect to all words $W$ in the $N$ most frequent words within the documents of a topic.

Now, for each pair $S_\beta$, we compute a confirmation measure $\phi(S_\beta)$~\citep{syed2017} as
\begin{align}
    \phi(S_\beta) = \frac{\boldsymbol{v}(W'_\beta)\cdot \boldsymbol{v}(W)}{||\boldsymbol{v}(W'_\beta)||\,||\boldsymbol{v}(W)||}.
\end{align}
The confirmation measure tells how strongly $W$ supports $W'$, i.e. how much semantically words from $W$ are related to $W'$ irrespective of how much two words (or sets of words) appear together in the corpus (see the work by~\citet{Roder:2015} for more detail on this). The average over all pairs $S_\beta$ are taken as the coherence for the specific topic under study. Further averaging over all topics, gives the $C_V$ coherence.

\end{document}